\documentclass[runningheads]{llncs}
\usepackage{graphicx}
\usepackage{multirow}
\usepackage{amssymb}
\usepackage{amsmath}
\usepackage{booktabs}

\begin{document}
\title{Textual Inversion and Self-supervised Refinement for Radiology Report Generation}
\titlerunning{Textual Inversion and Self-supervised Refinement for Report Generation}
%
\author{Yuanjiang Luo\inst{1}$^*$ \and
Hongxiang Li\inst{2}$^*$\and
Xuan Wu\inst{3}\and
Meng Cao\inst{4}\and
Xiaoshuang Huang\inst{5}\and
Zhihong Zhu\inst{2}\and
Peixi Liao\inst{6}\and
Hu chen\inst{3}$^\dagger$\and
Yi Zhang\inst{7}}
\authorrunning{Yuanjiang Luo  et al.}
%
\institute{National Key Laboratory of Fundamental Science on Synthetic Vision, Sichuan University, Sichuan, China
\and
Peking University, Shenzhen, China\\
\and
College of Computer Science, Sichuan University, Sichuan, China
\email{luoyj@stu.scu.edu.cn, huchen@scu.edu.cn}\\
\and
Mohamed bin Zayed University of Artificial Intelligence, United Arab Emirates
\and
China Agricultural University, Beijing, China
\and
The Sixth People's Hospital of Chengdu, Sichuan, China
\and
School of Cyber Science and Engineering, Sichuan University, Sichuan, China
}
\maketitle              
\def\thefootnote{$*$}\footnotetext{Equal contributions.}
\def\thefootnote{$\dagger$}\footnotetext{Corresponding author.}
\begin{abstract}
Existing mainstream approaches follow the encoder-decoder paradigm for generating radiology reports. They focus on improving the network structure of encoders and decoders, which leads to two shortcomings: overlooking the modality gap and ignoring report content constraints. In this paper, we proposed \textbf{T}extual \textbf{I}nversion and \textbf{S}elf-supervised \textbf{R}efinement \textbf{(TISR)} to address the above two issues. Specifically, textual inversion can project text and image into the same space by representing images as pseudo words to eliminate the cross-modeling gap. Subsequently, self-supervised refinement refines these pseudo words through contrastive loss computation between images and texts, enhancing the fidelity of generated reports to images. Notably, \textbf{TISR} is orthogonal to most existing methods, plug-and-play. We conduct experiments on two widely-used public datasets and achieve significant improvements on various baselines, which demonstrates the effectiveness and generalization of \textbf{TISR}. The code will be available soon.

\end{abstract}
\section{Introduction}

Radiology report generation provides the basis for physician diagnosis~\cite{ref-1}. However, observing radiograph and writing report is time-consuming and laborious for doctors~\cite{ref-2}. It's even error-prone for inexperienced doctors as they often struggle to accurately capture the abnormalities in images~\cite{ref-3,ref-4}. Previous approaches adopt the framework of image captioning~\cite{ref-6} straightforwardly and make it more suitable for generating radiology reports by improving image encoders~\cite{ref-7,ref-8,ref-9} to be better adapted to medical images or refining text decoders to generate long paragraphs~\cite{ref-11,ref-12,ref-13}. Building upon this, innovative techniques are utilized to improve performance, such as knowledge graph~\cite{ref-15}, causal inference~\cite{ref-18}, and dynamic graph~\cite{ref-19}.
\begin{figure}[t]
\centering
\includegraphics[width=1.\textwidth]{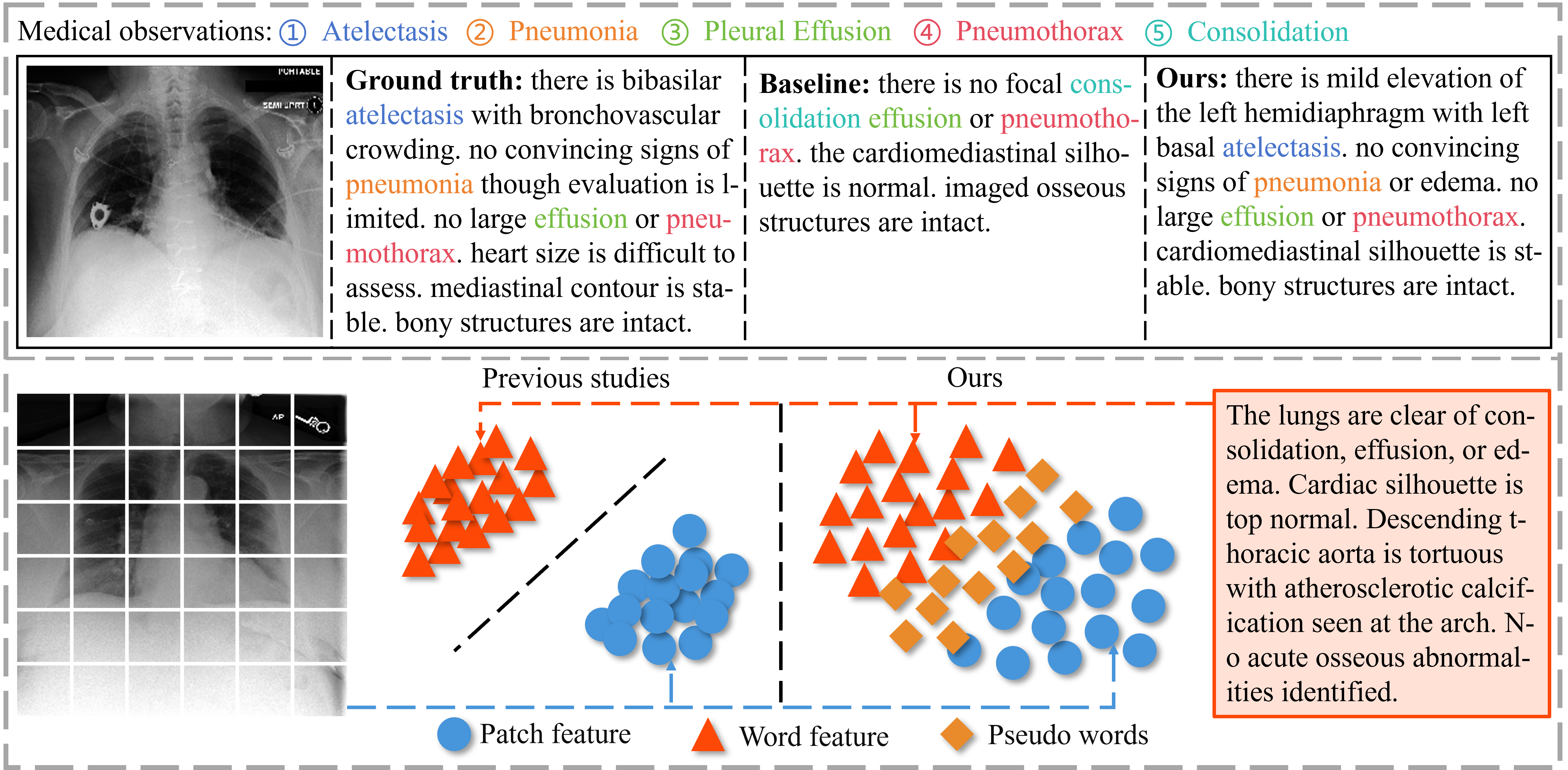}
\caption{Existing challenges in radiology report generation.} \label{fig1}
\end{figure}

Despite these notable advances, there are still two challenges in generating accurate reports. \textbf{(1)} Existing methods cannot explicitly constrain the reports generated by the text decoder to be faithful to visual information. Prior method~\cite{ref-20} suffers from the phenomenon of illusion generation, as shown in Fig.~\ref{fig1}. It generates ``\textit{consolidation}'', which is not mentioned in the ground truth while misses ``\textit{atelectasis}'' and ``\textit{pneumonia}''. Some works have enhanced the ability of grounding by extracting additional expert information, such as anchor box~\cite{ref-7} and sentence retrieval library~\cite{ref-21}. However, their implementation needs to make additional labels or reconstruct the entire dataset, which not only requires expensive costs but is not always accessible in clinical applications. \textbf{(2)} The inherent modal gap between images and language\cite{,ref-50}. Previous approaches~\cite{ref-7,ref-8,ref-9,ref-11,ref-12,ref-13,ref-14,ref-49,ref-51} adhere to the image encoder-text decoder paradigm~\cite{ref-22}, which lacks cross-modal interaction. As Image and text exist in distinct feature spaces with a feature gap between them, we propose to fill this gap with pseudo words, constructing a unified public hidden space for image and text, as shown in Fig.~\ref{fig1}.

We propose \textbf{T}extual \textbf{I}nversion and \textbf{S}elf-supervised \textbf{R}efinement \textbf{(TISR)} to solve the problems discussed above. We employ a lightweight mapping module, named textual inversion, to convert image features into text features~\cite{ref-25}. Through textual inversion, the pseudo words obtained by transforming image embeddings contain both image features and linguistic spatial characteristics. Textual inversion can eliminate the spatial gap effectively, making the features of two modalities be computed in the common compact space. We then perform self-supervised refinement by calculating contrastive loss between the obtained pseudo words and image features. Instead of relying on ground truth, \textbf{TISR} guides the network to generate reports faithful to the images by minimizing the contrastive loss\cite{ref-44}. Experimental results on two widely used datasets and three radiology report generation networks verify the efficacy and plug-and-play capability of \textbf{TISR}. In summary, the contributions of this paper are as follows:
\begin{itemize}
    \item We bridge the modality gap by transforming visual features into linguistic space through textual inversion.
\end{itemize}
\begin{itemize}
    \item The self-supervised refinement module searches for text representations close to the image content to minimize the contrastive loss. Consequently, we can generate faithful reports to radiographs, providing more credible diagnostic information for clinical practice.
\end{itemize}
\begin{itemize}
    \item Our \textbf{TISR} is orthogonal to other radiology report generation networks, plug-and-play. Experimental results show that by improving the network with \textbf{TISR}, the accuracy improved compared to the baselines.
\end{itemize}

\section{Method}
\begin{figure}[t]
\includegraphics[width=\textwidth]{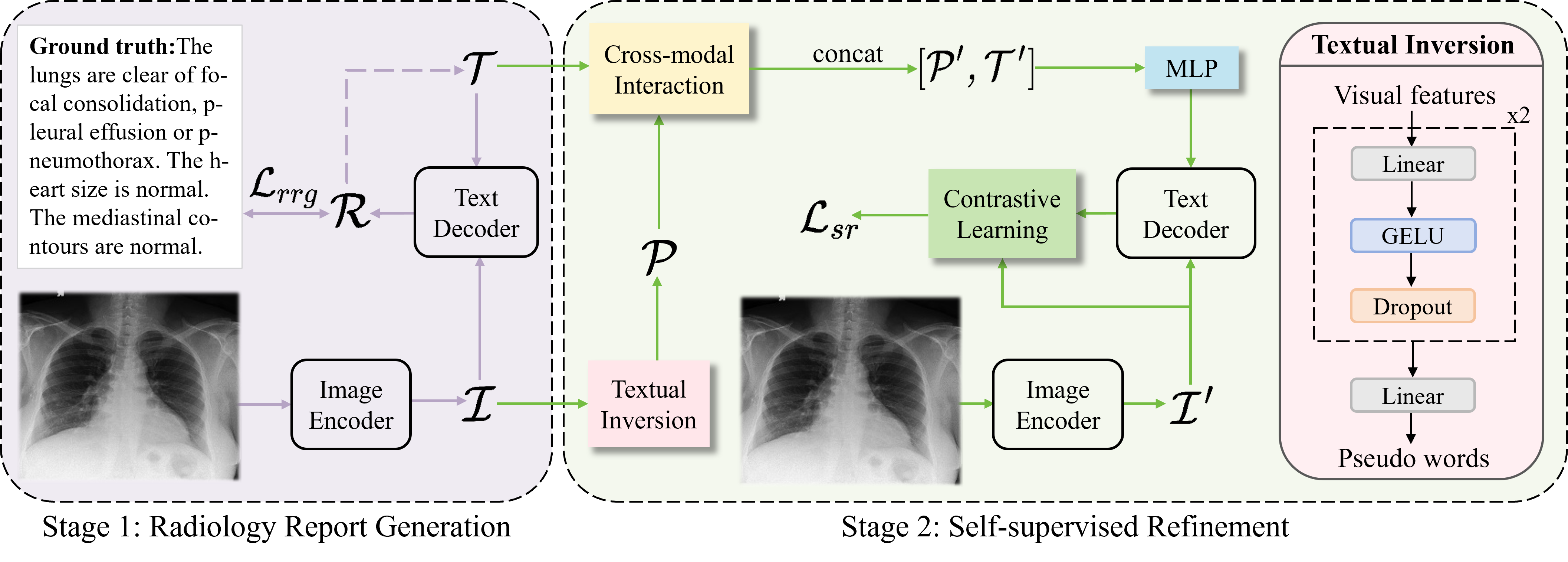}
\caption{Overview of our method. The arrow dashed line indicates that before obtaining the entire report, the current word is generated by the image features and text embeddings obtained by encoding the previously generated words. } \label{fig2}
\end{figure}
As shown in Fig.~\ref{fig2}, our pipeline consists of two stages. We extract image features $\mathcal{I} \in \mathbb{R}^{B\times M\times D}$ from a radiograph through an image encoder~\cite{ref-26}, where $B$ is the batch size, $M$ is the number of the processed patches and $D$ is the dimension. With the image features $\mathcal{I}$ and the previously generated text embeddings $\mathcal{T}_{n-1}$, the text decoder~\cite{ref-29} can obtain the log probability $\mathcal{O}_{n}$ of the next word.
\begin{equation}
\mathcal{O}_{n}=f_{d}(\mathcal{I},\mathcal{T}_{n-1})\quad  \text{and} \quad  \mathcal{T}_{n-1}=f_{e}(\mathcal{R}_{n-1}),
\end{equation}
where $f_{e}$, $f_{d}$ and n denote text encoder, text decoder and one word of the target report respectively. By applying linear and softmax to the log probability, we obtain the word in the vocabulary corresponding to the highest probability and use it as the n-th word of the report.
\begin{equation}
\mathcal{R}_{n}=\text{softmax}(f_{l}(\mathcal{O}_{n})).
\end{equation}
where $f_{l}$ refers linear function. We can finally obtain the complete text embeddings $\mathcal{T} \in \mathbb{R}^{B\times N\times D}$ and report $\mathcal{R} \in \mathbb{R}^{B\times N}$ after continuous autoregression of the text decoder~\cite{ref-30}, where $N$ represents the length of the target sequence.

Image features $\mathcal{I}$ are processed by textual inversion to generate pseudo words $\mathcal{P}$. In the self-supervised refinement, we supervise the network to generate more refined pseudo words by calculating the contrastive loss between text features and image features instead of using ground truth as the supervision signal. The details are illustrated in the following subsections.

\subsection{Textual Inversion}
Radiology report generation is an image-to-text cross-modal task, as medical images and radiology reports are in two different feature spaces. Existing methods are more tend to improve the overall performance by extracting refined image features~\cite{ref-7,ref-8,ref-9} or improving the network structure of the text decoder~\cite{ref-11,ref-12,ref-13,ref-14} and ignoring the gap between modalities. Therefore, we propose textual inversion to reconstruct image representation within the text embedding space to eliminate the spatial gap.
In this module, we map image embeddings $\mathcal{I}$ to pseudo words $\mathcal{P} \in \mathbb{R}^{B\times M\times D}$. via feeding image features into a three-layered full-connected network, which can be formulated as:
\begin{equation}
\mathcal{P} = f_{l}(\text{Dropout}(\text{GELU}(f_{l}(P_{1}))))\quad  \text{and} \quad
P_{1} = \text{Dropout}(\text{GELU}(f_{l}(\mathcal{I}))).
\end{equation}

\subsection{Self-supervised Refinement}
After obtaining the pseudo words, we input them into the text decoder after a series of operations to obtain $\mathcal{O'}$. We explicitly constrain that the generated pseudo words should be able to represent the image features sufficiently by calculating the contrastive loss between $\mathcal{O'}$ and the image feature $\mathcal{I'}$. This optimization process guides the network to generate reports that are faithful to images.

We first perform cross-modal interaction by employing cross attention mechanism~\cite{ref-31} between pseudo words $\mathcal{P}$ and text embeddings $\mathcal{T}$. Since pseudo words are derived directly from image features, it is beneficial to align visual and linguistic features through this interaction. This process can be expressed as:
\begin{equation}
\mathcal{P'}=\text{softmax}(\frac{\mathcal{P}\mathcal{T}^{T}}{\sqrt{D}})\mathcal{T},
\end{equation}
\begin{equation}
\mathcal{T'}=\text{softmax}(\frac{\mathcal{T}\mathcal{P}^{T}}{\sqrt{D}})\mathcal{P},
\end{equation}

We assume that the pseudo words can compensate for the missing information or correct the redundant information for the text feature $\mathcal{T}$. Based on this intuition, we concatenate aligned text feature $\mathcal{T'}$ with aligned pseudo words $\mathcal{P'}$. The concatenated features can be fused well through multi-layer perceptron (MLP), and thus we obtained the processed pseudo words $\mathcal{P''} \in \mathbb{R}^{B\times S\times D}$, where $S=M+N$. It can be expressed using a formula:
\begin{equation}
\mathcal{P''}=\text{MLP}[\mathcal{P'},\mathcal{T'}].
\end{equation}

The log probability $\mathcal{O'} \in \mathbb{R}^{B\times S\times D}$ is obtained by decoding $\mathcal{P''}$. We then implement self-supervised refinement by calculating contrastive loss between text embeddings $\mathcal{O'}$ and image features $\mathcal{I'}$. By minimizing the contrastive loss, we encourage the network to generate $\mathcal{P''}$ that closely resemble the expression of $\mathcal{I'}$. After continuous back-propagation and optimization, the generated pseudo words $\mathcal{P}$ can adequately represent the image semantics, which is beneficial for the generation of reports faithful to the original images. 

\subsection{Training Objective}
We utilize $\mathcal{L}_{rrg}$ to quantify the difference between the generated report and the ground truth~\cite{ref-33}, thus guiding the model to generate reports that are close to the ground truth. The formulation of $\mathcal{L}_{rrg}$ is as follows:
\begin{equation}
\mathcal{L}_{rrg}=- \frac{1}{\sum_{}^{}M}\sum_{b=1}^{B}\sum_{s=1}^{S}M_{bs}O_{bs,T_{bs}}.\label{eq:5}
\end{equation}

The log probability of the output against the target sequence $T$ is obtained by $O_{bs, T_{bs}}$ for the position $s$ of the $b$-th sample. To ensure consistent input sequence length, all sequences are filled to the same length. The mask $M_{bs}$ indicates whether a real word exists at the position: if present, it is 1; otherwise, it is 0. The log probability of the filled part is set to 0 by multiplying with the mask $M_{bs}$ to prevent the filled part from affecting $\mathcal{L}_{rrg}$. Finally, we normalize $\mathcal{L}_{rrg}$ by dividing it with the sum of all mask values $\sum_{}^{}M$ to ensure that the loss value is not affected by changes in sequence length.

In addition to optimizing the network to generate more accurate reports through $\mathcal{L}_{rrg}$, we also constrain the textual inversion to generate pseudo words that are close to the image representation through $\mathcal{L}_{sr}$. We obtained the score matrix $\mathcal{S}$ by calculating the correlation between image features and text features via dot product, which can be expressed as $\mathcal{S}=\mathcal{I'}\times \mathcal{O'}^{T}$, where $\times$ denotes matrix multiplication. we evaluate the cosine similarity between image features and text features and get the score matrix $\mathcal{S}$ of size $B \times B$. We optimize the network by constructing a symmetric cross-entropy loss to maximize the cosine similarity between $B$ real image-text pairs while minimizing the cosine similarity between $B^{2}-B$ unpaired image-text pairs~\cite{ref-34}.
\begin{equation}
\mathcal{L}_{sr}=-\frac{1}{2}(\sum_{b=1}^{B} \mathcal{M}_{b}log(S_{b})+\sum_{b=1}^{B} \mathcal{M}_{b}log({S_{b}}^{T})).
\end{equation}

$\mathcal{M}$ is a matrix of size $B \times B$, where the elements on the diagonal are 1, indicating positive samples, while the off-diagonal elements are 0, indicating negative samples. The overall loss function $\mathcal{L}$ of our network is defined as: $\mathcal{L} =\mathcal{L}_{rrg}+ \mathcal{L}_{sr}$. Instead of relying on manually labeled datasets, we leverage contrastive learning to measure the similarity between text and image, guiding the network to optimize parameters for generating reports faithful to the visual content.

\section{Experiment}
\subsection{Dataset and Evaluation Metrics}
\textbf{Dataset.} We conducted experiments on two widely-used datasets: the small dataset IU X-ray\footnote{https://openi.nlm.nih.gov/.}~\cite{ref-35} (containing 7,470 chest X-ray images and 3,955 corresponding reports) and the large dataset MIMIC-CXR\footnote{https://physionet.org/content/mimic-cxr/2.0.0/.}~\cite{ref-36} (containing 377,110 images and 227,835 corresponding reports). To ensure consistency and fairness in comparisons, we followed the data processing methods utilized by the three baselines~\cite{ref-18,ref-20,ref-33}. After excluding samples without corresponding radiology reports, IU X-ray is divided into training, validation and testing sets with a proportion of 7:1:2~\cite{ref-37} while MIMIC-CXR is divided according to the official splits~\cite{ref-33}.

\textbf{Evaluation Metrics.} We evaluate \textbf{TISR} on natural language generation (NLG) metrics including BLEU~\cite{ref-38}, METEOR~\cite{ref-39} and ROUGE-L~\cite{ref-40}, which are widely used to assess the fluency and accuracy of generated reports. We not only focus on the quality of the generated reports but also on their ability to accurately capture lesions in the images. Therefore, we employ clinical efficacy (CE) metrics to evaluate the detection accuracy of generated reports. CheXbert~\cite{ref-41} is applied to extract labels of 14 medical observations from reports. Precision, recall and F1 are calculated by comparing the labels of the generated reports with ground truth. 
\begin{table}[t]
\centering
\caption{Comparison between baselines and the improved network with \textbf{TISR}. $\triangle$ denotes the improvements compared to the baselines. * denotes our re-implementation of baselines. MTR and RG-L denote METEOR and ROUGE-L, respectively.}\label{tab1}
{\scriptsize
\begin{tabular}{l c c c c c c c c c}
\toprule
 \multirow{2}{*}{Method} & \multicolumn{6}{c}{NLG Metrics} & \multicolumn{3}{c}{CE Metrics} \\ \cmidrule(lr){2-7} \cmidrule(lr){8-10}
   & BLEU-1 & BLEU-2& BLEU-3& BLEU-4 & MTR & RG-L& Precision& Recall& F1\\
\midrule
\multicolumn{10}{c}{Experimental results on IU X-ray dataset.} \\
\midrule
R2Gen\textsuperscript{*}~\cite{ref-33}&0.443  &0.286  &0.212  &0.168  &0.175  &0.355&-&-&-  \\
  \textbf{+TISR(Ours)}&0.470  &0.310  &0.233  &0.187  &0.194  &0.369&-&-&-  \\
  $\triangle$  & \textbf{+0.027}  & \textbf{+0.024}  &\textbf{+0.021}  &\textbf{+0.019}  &\textbf{+0.019}  &\textbf{+0.014}&-&-&-  \\
 \hline
  R2GenCMN\textsuperscript{*}~\cite{ref-20}&0.469  &0.300  &0.215  &0.164  &0.190  &0.370&-&-&-  \\
  \textbf{+TISR(Ours)} &0.483  &0.313  &0.229  &0.176  &0.191  &0.371&-&-&-  \\
  $\triangle$  & \textbf{+0.014}  &\textbf{+0.013}  &\textbf{+0.014}  &\textbf{+0.012}  &\textbf{+0.001}  &\textbf{+0.001}&-&-&-  \\
  \hline
  VLCI\textsuperscript{*}~\cite{ref-18} & 0.467  &0.306  &0.225  &0.175  &0.193  &0.377&-&-&-  \\
  \textbf{+TISR(Ours)}&0.485  &0.318  &0.232  &0.179  &0.199  &0.382&-&-&-  \\
  $\triangle$ & \textbf{+0.018}  &\textbf{+0.012}  &\textbf{+0.007}  &\textbf{+0.004}  &\textbf{+0.006}  &\textbf{+0.005}&-&-&-  \\
\midrule
\multicolumn{10}{c}{Experimental results on MIMIC-CXR dataset.} \\
\midrule
R2Gen\textsuperscript{*}~\cite{ref-33}& 0.350&0.214&0.143  &0.103  &0.135  &0.271&0.424  &0.254  &0.317  \\
  \textbf{+TISR(Ours)}& 0.358&0.219&0.147  &0.106  &0.139  &0.275&0.467  &0.302  &0.367  \\
  $\triangle$     & \textbf{+0.008}& \textbf{+0.005}& \textbf{+0.004}   &\textbf{+0.003}  &\textbf{+0.004}  &\textbf{+0.004}&\textbf{+0.043}  &\textbf{+0.048}  &\textbf{+0.050}  \\
\midrule
  R2GenCMN\textsuperscript{*}~\cite{ref-20} & 0.344&0.210&0.139  &0.098  &0.136  &0.275&0.455  &0.317  &0.374  \\
  \textbf{+TISR(Ours)} & 0.363&0.224&0.149  &0.105  &0.143  &0.279&0.450  &0.344  &0.390  \\
  $\triangle$     & \textbf{+0.019} & \textbf{+0.014}& \textbf{+0.010}  &\textbf{+0.007}  &\textbf{+0.007}  &\textbf{+0.004}&-0.005  &\textbf{+0.027}  &\textbf{+0.016}  \\
\midrule
  VLCI\textsuperscript{*}~\cite{ref-18} & 0.393&0.239&0.159   &0.113  &0.150  &0.276&0.439  &0.283  &0.344  \\
 \textbf{+TISR(Ours)} & 0.396 &0.242&0.161 &0.115  &0.149  &0.278&0.453  &0.306  &0.366  \\
  $\triangle$     & \textbf{+0.003}& \textbf{+0.003}& \textbf{+0.002}  &\textbf{+0.002}  &\textbf{+0.001}  &\textbf{+0.002}&\textbf{+0.014}  &\textbf{+0.023}  &\textbf{+0.022}  \\
\bottomrule
\end{tabular}
}
\end{table}
\subsection{Experiments Results and Analyses}
\textbf{Comparison with Baselines.} To verify the generalization and effectiveness of \textbf{TISR}, we use R2Gen~\cite{ref-33}, R2GenCMN~\cite{ref-20} and VLCI~\cite{ref-18} as the baseline models in our experiments. These baseline models are improved with \textbf{TISR}, and the results are compared with the original baselines, as shown in Table~\ref{tab1}. Experimental results show that all metrics are enhanced by improving networks with \textbf{TISR}, which indicates that \textbf{TISR} can eliminate the gap between modalities and generate more accurate reports. It is remarkable that our approach does not require additional data and can seamlessly integrate into these baselines~\cite{ref-18,ref-20,ref-30}, which is of great importance for network migration and practical applications. What’s more, we can recognize from the results that prior methods have overlooked the impact of the gap between modalities on radiology report generation. Hence, future research should focus more on cross-modal interactions.

\textbf{Ablation Study.} To explore the effectiveness of each component in \textbf{TISR} and the rationality of the network structure, we conducted various ablation experiments. First of all, we explored the effectiveness of textual inversion and self-supervised refinement, as shown in Table~\ref{tab2}. The significance of textual inversion is investigated by calculating the contrastive loss between image features $\mathcal{I}$ and text embeddings $\mathcal{T}$, while the role of self-supervised refinement is explored through the calculation of the contrastive loss between image features $\mathcal{I}$ and pseudo words $\mathcal{P}$.
\begin{table}[t]
\centering
\caption{Ablation experiment of \textbf{TISR}.}\label{tab2}
{\scriptsize
\begin{tabular}{c c c c c c c c}
\toprule
\begin{tabular}[c]{@{}c@{}}Textual \\ Inversion\end{tabular}&\begin{tabular}[c]{@{}c@{}}Self-supervised  \\ Refinement\end{tabular}   &  BLEU-1 & BLEU-2 & BLEU-3 & BLEU-4 & METEOR & ROUGE-L\\
\midrule
 & &  0.443 &  0.286  & 
  0.212  & 0.168  &  0.175  &  0.355\\
 & \checkmark&  0.462  &  0.304
  &    0.224  & 0.176 &  0.190  &  0.361\\
\checkmark&  &  0.463  &  0.289
  &    0.206  & 0.157 &  0.181  &  0.356\\  
\midrule
 \checkmark& \checkmark&  \textbf{0.470}   &  \textbf{0.310}   &   \textbf{0.233}   & \textbf{0.187}  &  \textbf{0.194}   &  \textbf{0.369}\\
\bottomrule
\end{tabular}
}
\end{table}
\begin{table}[t]
\centering
\caption{Ablation experiments on the structure of textual inversion. }\label{tab3}
{\scriptsize
\begin{tabular}{ccccccc}
\toprule
  &BLEU-1 & BLEU-2 & BLEU-3 & BLEU-4 & METEOR & ROUGE-L\\
\midrule
Transformer &  0.416 &  0.268  &   0.194  & 0.149  &  0.178  &  0.350\\
\textbf{MLP} &  \textbf{0.470} &  \textbf{0.310}  &   \textbf{0.233}  &  \textbf{0.187}  &  \textbf{0.194}  &  \textbf{0.369}\\
\bottomrule
\end{tabular}
}
\end{table}

Secondly, we carried out experiments to explore the structure of \textbf{TISR}. We replace the three-layer linear structure with a three-layer transformer encoder to explore the structure of the textual inversion module. It’s easy to see that the result is worse than MLP with the same dimension of the hidden layer from Table~\ref{tab3}. We hypothesize that this is because the medical image features are not complex, so using a transformer may lead to overfitting, and it will also increase computational costs. We investigate the significance of cross-modal interaction and MLP by removing cross attention and MLP from the complete self-supervised refinement network respectively. Pseudo words' significance is investigated by directly incorporating $\mathcal{T}$ into the self-supervised refinement network. Furthermore, the importance of decoding text embeddings is explored by computing the contrastive loss between $\mathcal{P''}$ and $\mathcal{I'}$. We can speculate that each module plays an important role in generating more refined pseudo words from Table~\ref{tab4} since removing any one of them the performance of the network is degraded.
\begin{table}[t]
\centering
\caption{Ablation experiments on the structure of self-supervised refinement.}\label{tab4}
{\scriptsize
\begin{tabular}{cccccccccc}
\toprule
\begin{tabular}[c]{@{}c@{}}Text \\ Decoder\end{tabular}&\begin{tabular}[c]{@{}c@{}}Pseudo \\ Words\end{tabular} &\begin{tabular}[c]{@{}c@{}}Cross-modal\\ Interaction\end{tabular} &MLP  &  BLEU-1 & BLEU-2 & BLEU-3 & BLEU-4 & METEOR & ROUGE-L\\
\midrule
 & $\checkmark$& $\checkmark$& $\checkmark$&  0.442&  0.277 &  0.198  & 0.152&0.176&0.352\\
$\checkmark$& & $\checkmark$& $\checkmark$&0.446&  0.291 &  0.218  & 0.175&0.183&0.365\\
$\checkmark$&$\checkmark$& &  $\checkmark$&0.459&  0.298 &  0.221  & 0.173&0.186&0.358\\
$\checkmark$&$\checkmark$&  $\checkmark$& &0.432&  0.301 &  0.226  & 0.175&0.190&\textbf{0.398}\\
\midrule
$\checkmark$&$\checkmark$&  $\checkmark$&$\checkmark$ &\textbf{0.470}  &\textbf{0.310}  &\textbf{0.233}  &\textbf{0.187}  &\textbf{0.194}  &0.369  \\
\bottomrule
\end{tabular}
}
\end{table}

\begin{figure}[t]
\includegraphics[width=\textwidth]{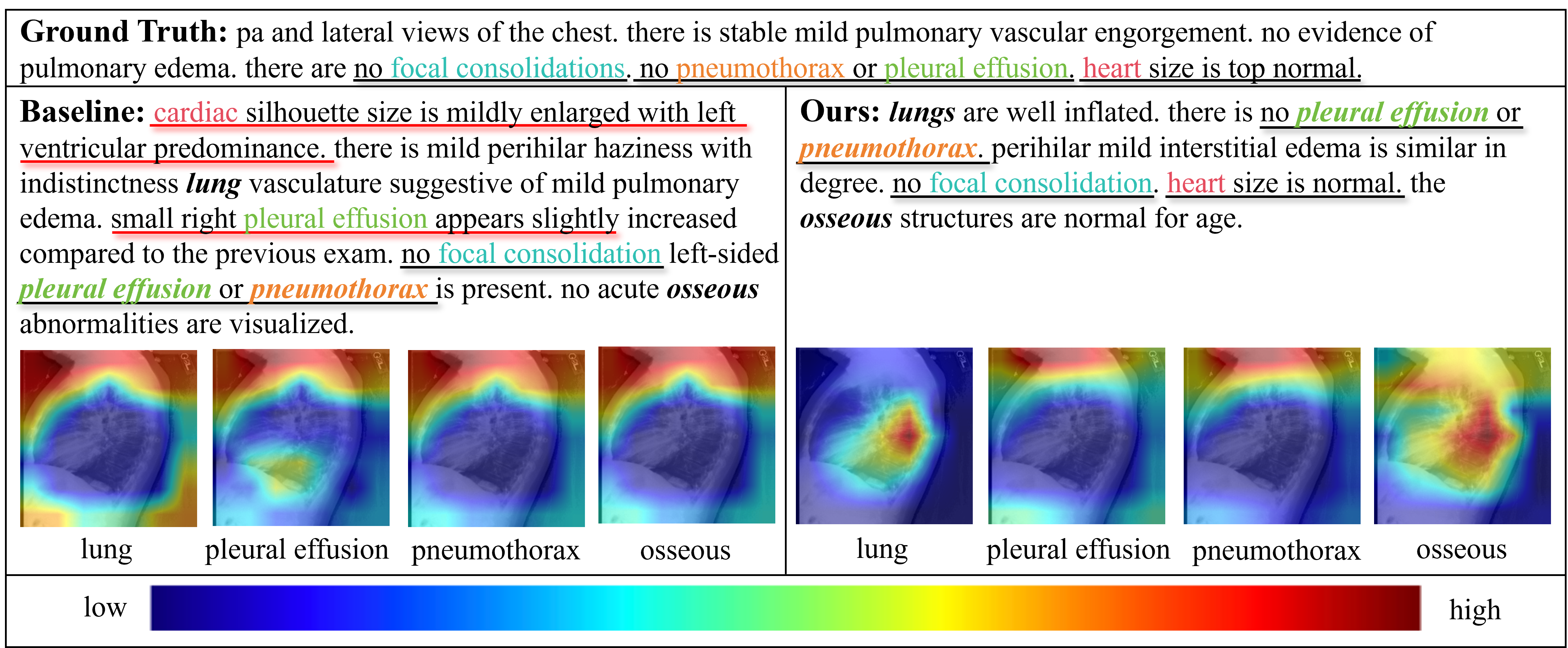}
\caption{Visualization. Red: the network is highly concerned about this area, blue: the area that is not concerned, black line: correct description, red line: incorrect description.} \label{fig3}
\end{figure}

\textbf{Quantitative Analysis.} We draw attention maps to explore the region of the medical image that the word of the generated report is interested in. Fig.~\ref{fig3} illustrates that the model improved by \textbf{TISR} is more sensitive to the correct regions and can generate reports that are closer to the ground truth. This demonstrates that our model can eliminate the cross-modal gap and thus generate reports faithful to images.
\section{Conclusion}
In this study, we propose \textbf{T}extual \textbf{I}nversion and \textbf{S}elf-supervised \textbf{R}efinement \textbf{(TISR)} to address the radiology report generation problem. By inverting image features into pseudo words, textual inversion aims to bridge the modality gap by representing visual features in the linguistic space. We employ self-supervised refinement to iteratively improve the quality of pseudo words by minimizing the contrastive loss between them and the image features. This iterative process helps to generate radiology reports that are faithful to the radiology image. \textbf{TISR} is designed to compensate for most existing approaches seamlessly, offering a plug-and-play solution. Significant improvements across all three baselines illustrate the effectiveness and generation of our proposed method.

%
%
%
%

\bibliographystyle{splncs04}
\bibliography{references}

\end{document}